\documentclass[10pt]{article}
\usepackage[preprint]{tmlr}

\usepackage{amsmath,amsfonts,bm}

\def\eqref#1{equation~\ref{#1}}

\def\1{\bm{1}}

\DeclareMathAlphabet{\mathsfit}{\encodingdefault}{\sfdefault}{m}{sl}
\SetMathAlphabet{\mathsfit}{bold}{\encodingdefault}{\sfdefault}{bx}{n}

\newcommand{\R}{\mathbb{R}}

\usepackage{amsmath,amssymb,amsthm,mathtools}
\usepackage{booktabs}
\usepackage{graphicx}
\usepackage{float}
\usepackage{microtype}
\usepackage{xcolor}
\usepackage{hyperref}
\usepackage{url}
\hypersetup{
    hidelinks,
    pdfauthor={Jiming Feng and Junliang Li},
    pdftitle={Scaled Idempotence in Transformer Attention: Paired OV Geometry and Shared-Value Algebras}
}
\author{
\name Jiming Feng \email jming2011@emails.bjut.edu.cn \\
\addr College of Computer Science, Beijing University of Technology \\
Beijing, China
\AND
\name Junliang Li \\
\addr School of Mathematics, Statistics and Mechanics, Beijing University of Technology \\
Beijing, China
}

\title{Scaled Idempotence in Transformer Attention:\\Paired OV Geometry and Shared-Value Algebras}

\newcommand{\Pscore}{\mathcal{P}}

\newtheorem{proposition}{Proposition}

\begin{document}

\maketitle

\begin{abstract}
We identify a recurrent algebraic regularity in Transformer attention: a sparse subset of effective OV operators $T=OV^\top$ nearly closes under composition, $T^2\approx\alpha T$.\footnote{Large language model tools assisted with code drafting, experimental-design iteration, analysis scripting, literature discovery, and language editing. The research question and final scientific judgments are the authors'; all derivations, executions, numerical claims, citations, and final text were independently checked, and the authors assume full responsibility.} Across six pretrained endpoints spanning 2.8B--235B parameters, 3.98--8.00\% of heads reach squared closure alignment $\Pscore\geq0.9$, whereas no matched within-layer O/V mismatch reaches this threshold. The tail is not numerically square-zero: over all 881 strong heads, the return gain $\lVert T^2\rVert_F/\lVert T\rVert_F$ has model medians 0.032--2.51 and a pooled minimum of 0.014. An exact factorization in principal coordinates,
\[
T=Q_OKQ_V^\top,\qquad T^2=Q_O(KDK)Q_V^\top,
\]
separates within-support transport $K$ from read--write return geometry $D$. A population-wide intervention over all 7,304 heads in nine MHA/GQA models scrambles only the orientation of $K$ while preserving its singular spectrum, norm, read/write factor spans, and every principal angle; all 14,608 participating O/V factors have full numerical column rank. Median closure falls from 0.336 to $1.04\times10^{-4}$; the trained orientation yields higher closure for 98.64\% of heads and in all 268 layers. A complementary search, with heads sampled independently of closure, finds an explicit feasible construction above 0.8 among the sampled heads in every layer, including layers without a strong head, although most sampled high-capacity heads do not attain strong closure. Retrospective trajectories in three independently trained lineages further separate broadly available feasible geometry from the orientations attained by the final strong population. Under exact value sharing, headwise closure extends to a fixed-gain right-action algebra: $T_iT_j=\alpha_jT_i$ for every sibling $i$. Seven-model experiments verify the approximate law and reveal distinct oblique projections with a shared value-defined kernel. Together, these results characterize scaled idempotence as a sparse trained orientation within a broadly available geometric capacity and show that value sharing extends the headwise relation to a local operator algebra.
\end{abstract}

\section{Introduction}\leavevmode\par

For a generic low-rank map, applying it twice need not preserve the direction of applying it once. Yet a nontrivial minority of trained Transformer attention heads exhibit
\begin{equation}
    T^2 \approx \alpha T, \qquad T=OV^\top, \label{eq:scaled-idempotence}
\end{equation}
where $T$ is the head's effective residual-stream OV operator. This scaled idempotence is visually and algebraically suggestive: the operator nearly closes under composition. It also poses a concrete mechanistic question. Is Equation~\ref{eq:scaled-idempotence} a chance consequence of low rank, an approximate projection or copying rule, or the visible endpoint of a relation learned between the separately parameterized value and output factors?

To distinguish these possibilities, we compare each canonical O/V pair with cyclic within-layer wrong-V pairings that preserve the two factor collections while breaking their trained correspondence. Strong scaled idempotence disappears under this control. We then decompose each operator into principal-angle geometry and a transport core, allowing the contribution of their head-specific pairing to be tested directly.

Our results attribute Equation~\ref{eq:scaled-idempotence} to a specific relation between separately parameterized O and V factors and establish an exact compositional consequence under value sharing. The relation does not require $T$ to be an orthogonal projection, the read and write subspaces to coincide, or the head to copy tokens. We derive a small-matrix factorization that separates subspace support from within-subspace transport. Let the principal-angle cosines between the read and write spaces form a diagonal matrix $D$, and let $K$ express the OV core in the corresponding principal coordinates. Then
\begin{equation}
    T = Q_O K Q_V^\top, \qquad T^2 = Q_O(KDK)Q_V^\top. \label{eq:kdk-intro}
\end{equation}
When $D\approx cI$, closure of $T$ is governed by the more general latent law $K^2\approx\beta K$; the transport core need not be close to a scalar matrix. More importantly, the decomposition makes the mechanism experimentally separable: support geometry and transport orientation can be intervened on while their coarse invariants remain fixed.

We study this phenomenon across model scale, attention topology, training stage, and matched geometric controls. Our contributions are:
\begin{itemize}
    \item We identify a recurrent scaled-idempotent upper tail across pretrained Transformer endpoints from 2.8B to 235B parameters, show that it disappears under marginal-preserving O/V mismatch, and derive its exact $KDK$ return geometry.
    \item We isolate transport orientation without selecting heads by closure. Across all heads and layers in nine MHA/GQA models, closure collapses when the trained orientation is scrambled while spectrum, read/write spans, and principal angles remain fixed.
    \item We separate geometric capacity from trained attainment. High closure is constructively feasible in every surveyed layer, while retrospective trajectories associate the sparse final tail primarily with attained orientation rather than with scarce capacity.
    \item We prove that shared values turn exact headwise closure into a fixed-gain cross-head operator algebra. Seven-model tests confirm the approximate law, and an affine normal form characterizes distinct oblique projections with a common value-defined kernel.
\end{itemize}

Scaled idempotence is therefore a sparse signature of a paired O/V relation, not a property visible in either factor alone.

\section{Related Work}

Mechanistic interpretability commonly separates QK routing from OV content transformation~\citep{elhage2021mathematical}; its V-composition scores already ask whether one component writes into directions read by another. Subsequent work has identified task-local OV circuits for induction, indirect-object identification, and algorithmic computation~\citep{olsson2022context,wang2023interpretability,nanda2023progress}. The recent \emph{Communication Map} extends operator-product analysis into a task-free census of component-to-component communication strength~\citep{wang2026communication}. Our object is different: whether one learned OV operator closes \emph{directionally} under reapplication, and whether value sharing turns that self-relation into a fixed-gain cross-head law.

Our analysis uses principal angles and Grassmannian subspace similarity~\citep{bjorck1973numerical,hamm2008grassmann}. Recent work compares attention-head weight subspaces in GPT-2~\citep{yamagiwa2026affinity}, while \citet{chen2026multiplication} find projection-like OV routing in a one-layer modular-multiplication transformer. Principal-angle overlap discards the signed transport orientation retained by $K$. Our matched intervention holds the complete angle spectrum fixed while varying this additional degree of freedom.

Idempotence has also been imposed explicitly on neural networks~\citep{jensen2025enforcing}, and scaled-idempotent projection bases appear in theoretical analyses of in-context learning~\citep{bu2024provably}. In contrast, the relation studied here emerges without an idempotence objective in pretrained language-model OV weights, permits signed $\alpha$, and is measured by direct self-composition. GQA provides the architectural premise of shared key--value heads~\citep{ainslie2023gqa}; we derive and test the resulting right-action law. The present work connects a recurrent scaled-idempotent OV relation to its $KDK$ paired-return geometry and its algebraic extension under value sharing.

\section{Scaled Idempotence and Its Paired-Return Decomposition}
\label{sec:method}

\subsection{Effective OV operator}

Ignoring token-to-token attention mixing, write the residual-stream map of head $h$ as
\begin{equation}
    T_h = O_h V_{g(h)}^\top, \qquad O_h,V_{g(h)}\in\R^{d\times r}, \label{eq:ov}
\end{equation}
where $d$ is the model width, $r$ the head dimension, and $g(h)$ selects the value head. In MHA, each query head has its own value head; in GQA, several query/output heads share one value factor. The factorization is gauge-dependent, but $T_h$ is not.

The small cross-factor matrix
\begin{equation}
    M_h=V_{g(h)}^\top O_h \label{eq:return-core}
\end{equation}
is the return core of the pairing: $T_h^2=O_hM_hV_{g(h)}^\top$. It records how directions written through $O_h$ are read back through the value factor. Our primary object is the relation represented by $M_h$ and its basis-invariant support geometry, rather than either factor in isolation.

\subsection{Scaled idempotence as a composite endpoint}

We measure scaled idempotence as
\begin{align}
 \Pscore(T)
 &= 1-\frac{\min_{\alpha}\|T^2-\alpha T\|_F^2}{\|T^2\|_F^2} \\
 &= \frac{\langle T^2,T\rangle_F^2}{\|T^2\|_F^2\|T\|_F^2},
 \qquad
 \alpha^*=\frac{\langle T^2,T\rangle_F}{\|T\|_F^2}. \label{eq:p-score}
\end{align}
Thus $\Pscore$ is a squared cosine in operator space. It lies in $[0,1]$ and is invariant to uniform rescaling of $T$. All computations use $r\times r$ Gram matrices; no $d\times d$ operator need be materialized.
We define the score only when both $T$ and $T^2$ are nonzero; every empirical operator in the reported analyses satisfies this condition. Because the square removes the sign of proportionality, $\Pscore$ measures projective closure and permits either positive or negative fitted $\alpha$.

\subsection{Principal coordinates and the KDK identity}

Take reduced QR factorizations $O=Q_OR_O$ and $V=Q_VR_V$. Let
\begin{equation}
    Q_O^\top Q_V = ADB^\top,
    \qquad D=\operatorname{diag}(\cos\theta_1,\ldots,\cos\theta_r)
\end{equation}
be an SVD. Define $\widetilde Q_O=Q_OA$, $\widetilde Q_V=Q_VB$, and
\begin{equation}
    K=A^\top R_OR_V^\top B.
\end{equation}
Then $\widetilde Q_V^\top\widetilde Q_O=D$, giving the exact identities
\begin{equation}
    T=\widetilde Q_OK\widetilde Q_V^\top,
    \qquad
    T^2=\widetilde Q_O(KDK)\widetilde Q_V^\top. \label{eq:kdk}
\end{equation}
Computing Equation~\ref{eq:p-score} from $(K,D)$ exactly reconstructs the direct low-rank score for all analyzed heads. We therefore write the same quantity in principal coordinates as
\begin{equation}
 \mathcal P_D(K)=
 \frac{\langle KDK,K\rangle_F^2}
 {\lVert KDK\rVert_F^2\lVert K\rVert_F^2}
 =\Pscore(T). \label{eq:principal-closure}
\end{equation}
Two support summaries will be useful. Their total overlap is
\begin{equation}
 G=\frac{\lVert D\rVert_F^2}{r}
   =\frac{1}{r}\sum_i\cos^2\theta_i, \label{eq:support-overlap}
\end{equation}
and their scale-free isotropy is
\begin{equation}
 I_D=\frac{r\bar c^2}{\sum_i c_i^2},
 \qquad \bar c=\frac{1}{r}\sum_i c_i.
\end{equation}
For independent random $r$-dimensional subspaces in $\R^d$, $\mathbb E[G]=r/d$; neither $G$ nor $I_D$ records the signed orientation carried by $K$.

The latent transport law is measured directly by
\begin{equation}
 \beta_K^*=\frac{\langle K^2,K\rangle_F}{\lVert K\rVert_F^2},
 \qquad \mathcal P_I(K)=
 \frac{\langle K^2,K\rangle_F^2}
 {\lVert K^2\rVert_F^2\lVert K\rVert_F^2}. \label{eq:latent-closure}
\end{equation}
If $D=cI$, then $\Pscore(T)=\mathcal P_I(K)$ and the optimal outer coefficient is $\alpha_T^*=c\beta_K^*$. Thus isoclinic support exposes the general scaled-idempotent law $K^2\approx\beta K$; it does not require $K$ to resemble a scalar matrix. The coefficient $c$ may be far below one, so a map between distinct isoclinic subspaces can close exactly without being an identity or orthogonal projection.

Non-isoclinic support admits an exact perturbation description. Assume $K$, $K^2$, and $KDK$ are nonzero. Set $c=\operatorname{tr}(D)/r>0$, write $D=cI+E$, and define $X=K^2$, $Y=K$, and $F=KEK$. Let $x=X/\lVert X\rVert_F$, set $\sigma=\operatorname{sign}\langle X,Y\rangle_F$ (with $\sigma=1$ when the inner product is zero), and define $y=\sigma Y/\lVert Y\rVert_F$ and $s=\langle x,y\rangle_F=\sqrt{\mathcal P_I(K)}$. For $s<1$, let $u=(y-sx)/\sqrt{1-s^2}$ and decompose
\begin{equation}
 \frac{F}{c\lVert X\rVert_F}=ax+bu+w,
 \qquad w\perp x,u.
\end{equation}
Direct substitution gives
\begin{equation}
 \mathcal P_D(K)=
 \frac{\left[s(1+a)+\sqrt{1-s^2}\,b\right]^2}
 {(1+a)^2+b^2+\lVert w\rVert_F^2}. \label{eq:perturbation-identity}
\end{equation}
The radial term $a$ cancels when acting alone, $b$ rotates within the closure-relevant plane, and $w$ enters only the denominator as pure closure leakage. When $s=1$, the corresponding one-dimensional decomposition gives $(1+a)^2/[(1+a)^2+\lVert w\rVert_F^2]$ whenever $KDK\neq0$. Bounds and numerical certificates derived from this identity are reported in the supplement.

The same law also has a direct factor characterization. With $M=V^\top O$,
\begin{equation}
 T^2-\alpha T=O(M-\alpha I)V^\top,
 \qquad
 T^2=\alpha T\Longleftrightarrow V^\top O=\alpha I, \label{eq:cross-gram-equivalence}
\end{equation}
where the equivalence holds for full-column-rank factors. Under the gauge $O\mapsto OG$, $V\mapsto VG^{-\top}$, $M$ changes by similarity, so the scalar-identity condition is gauge invariant.

Assuming full-column-rank $O,V$, the approximate statement is exact in the factor-induced metric. Let $A=O^\top O$, $B=V^\top V$, and $\langle X,Y\rangle_{A,B}=\operatorname{tr}(X^\top A YB)$. Then
\begin{equation}
 \langle OXV^\top,OYV^\top\rangle_F=\langle X,Y\rangle_{A,B},
 \qquad
 \Pscore(T)=\frac{\langle M,I\rangle_{A,B}^2}{\lVert M\rVert_{A,B}^2\lVert I\rVert_{A,B}^2}. \label{eq:weighted-crossgram}
\end{equation}
The fitted coefficient is $\alpha^*=\langle M,I\rangle_{A,B}/\lVert I\rVert_{A,B}^2$. Thus $T^2\approx\alpha T$ is equivalent to weighted projective scalarity of the paired return core $M$. Ordinary unweighted Frobenius scalarity additionally depends on factor conditioning and is not implied uniformly.

\begin{proposition}[Shared-value operator algebra]
\label{prop:shared-value-algebra}
Let $V\in\mathbb R^{d\times r}$ have full column rank and define $C_V=V(V^\top V)^{-1}$. The family
\begin{equation}
 \mathcal A_V=\left\{X_{a,N}=(aC_V+N)V^\top:
 a\in\mathbb R,\;V^\top N=0\right\} \label{eq:shared-value-algebra}
\end{equation}
is a linear operator algebra satisfying
\begin{equation}
 X_{a,N}X_{b,M}=bX_{a,N}. \label{eq:algebra-product}
\end{equation}
Consequently $X_{a,N}^2=aX_{a,N}$. Conversely, if $T=OV^\top$ with $O,V$ full column rank, then $T^2=aT$ if and only if $T=X_{a,N}$ for some $N$ with $V^\top N=0$.
\end{proposition}

The product follows immediately from $V^\top C_V=I$ and $V^\top N=0$. For $a=1$, the normalized slice consists of oblique projections $P_N=(C_V+N)V^\top$ with common kernel $\ker(V^\top)$ and $P_NP_M=P_N$. When $r<d$, $\mathcal A_V$ is nonunital: value sharing defines a local algebra rather than a copy of the full residual-stream algebra.

The exact law also controls approximation. For shared-$V$ heads $T_i=O_iV^\top$ and $T_j=O_jV^\top$, let $A_i=O_i^\top O_i$, $A_j=O_j^\top O_j$, and $\gamma_{ij}=\lambda_{\max}(A_j^{-1/2}A_iA_j^{-1/2})$. If $O_j$ has full column rank, then for every $\alpha$,
\begin{equation}
 \lVert T_iT_j-\alpha T_i\rVert_F^2
 \leq \gamma_{ij}\lVert T_j^2-\alpha T_j\rVert_F^2. \label{eq:shared-value-bound}
\end{equation}
Thus exact closure of the right head propagates its coefficient to every sibling, while the approximate law can be amplified by relative output-factor conditioning and must be measured empirically. A proof of Equation~\ref{eq:shared-value-bound} is given in the supplement.

\section{Experimental Design}

Each experiment addresses a distinct question and therefore uses a dedicated model population, summarized in Table~\ref{tab:study-map}. The six-endpoint survey provides broad scale coverage; the nine-model analysis uses complete factors for matched interventions; and the remaining experiments examine capacity, endpoint geometry, training trajectories, and the algebraic consequences of value sharing.

\begin{table}[H]
\caption{Study map. Each population has one primary inferential role.}
\label{tab:study-map}
\centering
\small
\resizebox{\linewidth}{!}{%
\begin{tabular}{lll}
\toprule
Question & Population & Primary comparison \\
\midrule
Does closure recur? & 6 endpoints; 15,936 all heads & canonical versus wrong-$V$ \\
What degree of freedom carries it? & 9 models; 7,304 all heads & fixed-invariant orientation scramble \\
Why is the upper tail sparse? & 268 layers plus 3 trajectories & feasible capacity versus attained orientation \\
What geometry survives normalization? & 12 endpoints; 1,180 strong heads & latent core versus trained support \\
What does value sharing imply? & 7 GQA models; 1,458 products & fixed-gain sibling composition \\
\bottomrule
\end{tabular}}
\end{table}

\subsection{Cross-model survey without head selection}

We scan every attention head in six fully trained endpoints: Pythia-2.8B-deduped~\citep{biderman2023pythia}, Qwen3-4B Base~\citep{yang2025qwen3}, Mistral-7B-v0.3~\citep{jiang2023mistral}, OLMo-2-13B~\citep{walsh2025olmo2}, Qwen2.5-72B Base~\citep{qwen2024qwen25}, and Qwen3-235B-A22B. This covers 15,936 heads, MHA and GQA, dense and MoE architectures. For each head, a cyclic wrong-$V$ control preserves the layerwise factor collections while breaking trained identity.

All quantities are evaluated through exact $r\times r$ identities in float64 and checked against direct calculations. We use $\Pscore\geq0.9$ as a fixed descriptive convention and sweep thresholds from 0.75 to 0.975 over all six populations. Because $1-\Pscore$ is the minimum normalized squared residual after fitting $T^2\approx\alpha T$, the threshold corresponds to a relative residual of at most $\sqrt{0.1}\approx0.316$.

\subsection{Factorial test of the KDK mechanism}

The factorial experiment uses every head in six independent families: Pythia-2.8B-deduped and OLMo-1B~\citep{groeneveld2024olmo} for MHA; TinyLlama-1.1B~\citep{zhang2024tinyllama}, SmolLM3-3B~\citep{bakouch2025smollm3}, Qwen3-4B Base, and Mistral-7B for GQA. Within each layer, we exhaustively reassign $K$ and $D$ separately. GQA donors move by whole KV groups, changing the value factor while holding the query slot fixed. The primary endpoint is the layer mean of
\begin{equation}
 I_{DK}=\mathcal P_{D_h}(K_h)-\mathcal P_{D_h}(K_{h'})
 -\mathcal P_{D_{h''}}(K_h)+\mathcal P_{D_{h''}}(K_{h'}). \label{eq:kdk-interaction}
\end{equation}
Layers are the inferential units. As a matched control, every $O_h$ receives a wrong $V$ from another head or KV group; $K,D$ are recomputed before repeating the factorial test. The complete correction and admission protocol is given in the supplement.

\subsection{Spectrum- and support-preserving intervention}

The factorial test changes the donor core and therefore does not hold its singular spectrum fixed. We construct a stricter counterfactual for every head. In principal coordinates, replace
\begin{equation}
 K \longmapsto K'=P_LKP_R^\top, \label{eq:spectral-intervention}
\end{equation}
where $P_L,P_R$ are independently sampled signed permutation matrices. Because they are orthogonal, $K'$ has exactly the same singular values as $K$. Keeping $Q_O,Q_V$, and $D$ fixed also preserves rank, every unitarily invariant norm, the read/write factor spans, and their complete principal-angle spectrum. An all-head audit confirms full numerical column rank for all 14,608 participating O/V factors under the tolerance $\max(m,n)\epsilon_{64}s_{\max}$; the smallest observed $s_{\min}/s_{\max}$ is $2.63\times10^{-4}$. For GQA, we refactor $Q_OK'Q_V^\top=O'V^\top$ using the original shared value factor, so group topology is unchanged. Eight deterministic interventions are applied to all 7,304 heads across seven GQA and two MHA families.

Repeated or nearly repeated principal angles make individual principal vectors non-unique. We therefore add an audit whose null distribution is basis invariant: $P_L$ and $P_R$ are replaced by independent Haar orthogonal matrices. Haar measure is unchanged by any orthogonal reparameterization within a degenerate principal-angle block. Two deterministic draws are evaluated per head, while the complete singular spectrum and support geometry remain fixed.

To separate geometric capacity from trained attainment, we hold the same invariants fixed and consider
\begin{equation}
 \operatorname{Cap}(\sigma,D)=
 \max_{U_L,U_R\in\mathsf O(r)}
 \mathcal P_D\!\left(U_L\operatorname{diag}(\sigma)U_R^\top\right). \label{eq:conditional-capacity}
\end{equation}
Here $\sigma$ is the trained singular spectrum. Projected ascent with exact SVD retraction provides a constructive lower bound $\widehat{\operatorname{Cap}}\leq\operatorname{Cap}$. We evaluate all 510 strong heads and deterministic layer-matched ordinary controls. A complementary sample selects two heads per layer by a fixed hash rule, independently of closure, across all 268 layers, including those without a strong head. Model- and layer-level conclusions resample models and complete layers; numerical-route and rank-one boundary audits are reported in the supplement.

\subsection{Supporting natural-activation diagnostic}

To test whether the weight-space law survives the anisotropic distribution of real hidden states, we use four deterministic 256-token WikiText segments in OLMo-1B, TinyLlama-1.1B, SmolLM3-3B, and Qwen3-4B-Instruct. The high-closure decile is paired with an equal number of low-closure heads from the same eligible layers. Writing activations as columns, if $y$ is a head's natural post-$O$ output, the endpoint is the squared directional alignment of $y$ with its diagnostic return $Ty=OV^\top y$. We also record the RMS return magnitude $g_y=(\sum\lVert Ty\rVert_2^2/\sum\lVert y\rVert_2^2)^{1/2}$. A cyclic wrong-$V$ pairing holds $y$ and $O$ fixed while breaking the trained return factor. The fixed population contains 271 high/low pairs (542 heads) across 86 eligible layers; this is a data-weighted consistency check, not a forward-circuit intervention.

\subsection{Latent-core and support decomposition}

Since $\mathcal P_{\lambda D}(K)=\mathcal P_D(K)$, total support magnitude cannot directly change closure. We replace trained support by $D_{\mathrm{iso}}=(\lVert D\rVert_F/\sqrt r)I$, preserving $K$ and total overlap while exposing the latent score $\mathcal P_I(K)$. We apply this normalization to 1,180 strong heads across twelve endpoints and test specificity against all 7,304 local heads plus layer-matched ordinary heads from the three large endpoints.

Conditional on latent closure, $I_D$ tests whether support shape preserves the core law. Equation~\ref{eq:perturbation-identity} then separates radial response, in-plane rotation, and off-plane leakage. The main text treats this analysis as a decomposition of endpoint geometry rather than a causal account of formation; threshold-sweep and clustered-inference summaries, axis-assignment controls, and perturbation bounds are reported in the supplement.

\subsection{Capacity and attainment across training}

We track fixed final strong/ordinary populations through independently pretrained OLMo-1B, TinyLlama-1.1B, and Pythia-2.8B-deduped trajectories. At four representative checkpoints per lineage, we recompute feasible capacity under the same spectrum/support constraints. If $R$ is the median matched-scramble closure, the constructed-range attainment index is
\begin{equation}
 E=\frac{\Pscore-R}{\widehat{\operatorname{Cap}}-R}. \label{eq:attained-fraction}
\end{equation}
Values are not clipped. Because the denominator uses a constructive lower bound rather than a certified global optimum, $E$ is a diagnostic index, not a certified fraction of total capacity. This retrospective comparison asks whether training creates a scarce feasible set or increasingly realizes a compatible orientation within an already capable set; it does not identify the examples or gradients that select final membership.

\subsection{Shared-value composition and gain-inheritance tests}

Proposition~\ref{prop:shared-value-algebra} predicts that an exactly closed right head acts with its own coefficient on every sibling sharing $V_g$. We test the approximate law in seven independently trained GQA families: TinyLlama-1.1B, SmolLM3-3B, Qwen3-4B Base, Mistral-7B, Granite-3.3-2B Base~\citep{granite33model}, Falcon3-3B Base~\citep{falcon3model}, and Yi-1.5-6B Base~\citep{yi15model}. Right heads satisfy the common threshold $\Pscore(T_j)\geq0.9$, and every distinct sibling $i\neq j$ is evaluated. The matched control uses the same query slot in the next KV group, preserving layer and architecture while breaking shared-$V$ identity.

Direction alone permits a different coefficient for every product, so we also test whether the right head transmits its own gain. For each pair, define
\begin{equation}
 F_{ij}(\alpha_j)=1-\frac{\|T_iT_j-\alpha_jT_i\|_F^2}{\|T_iT_j\|_F^2},\quad
 \beta_{ij}=\frac{\langle T_iT_j,T_i\rangle_F}{\|T_i\|_F^2},\quad
 \rho_{ij}=\frac{\beta_{ij}}{\alpha_j}. \label{eq:gain-inheritance}
\end{equation}
Here $\alpha_j$ is fitted once from $T_j^2\approx\alpha_jT_j$ and is not refitted to a sibling. After fixing the coefficient and control definitions on the development families, we evaluate Yi-1.5-6B as an additional-model confirmation. We also test whether normalized strong siblings $P_h=T_h/\alpha_h$ are distinct operators satisfying $P_iP_j\approx P_i$ rather than duplicate maps.

For the affine-normal-form prediction, we compare each normalized strong head with its canonical value-defined anchor and a cyclic wrong-$V$ anchor, then rank every compatible value factor in the same layer for seven GQA and two MHA models with complete candidate sets. OLMo-2-13B, Qwen2.5-72B Base, and Qwen3-235B-A22B provide an additional-model extension to larger scales.

\section{Results}

\subsection{Scaled idempotence recurs from 2.8B to 235B parameters}

Under a common closure statistic, threshold, and marginal-preserving wrong-$V$ control, all six fully trained endpoints contain a continuous upper tail with $\Pscore\geq0.9$ (Table~\ref{tab:closure-survey}). The tail contains 3.98--8.00\% of heads, or 67--338 heads per model. No wrong-$V$ control reaches the threshold, and control medians range from $5.7\times10^{-5}$ to $1.3\times10^{-4}$. Because both canonical and mismatched maps have rank at most $r$, low rank alone cannot account for the separation.

The tail persists throughout the fixed 0.75--0.975 threshold sweep, whereas no mismatch reaches even the lowest tested threshold. The marginal-preserving wrong-$V$ control supplies the pairing-specific comparison used uniformly across all six endpoints.

Across the 13B--235B endpoints, 69--76\% of high-closure heads have negative fitted $\alpha$. The recurring structure is therefore a signed return relation rather than only positive ``copy and amplify.''

\begin{table}[H]
\caption{Unified scaled-idempotence survey. All endpoints use the same score $\Pscore$ and strong-head threshold $\Pscore\geq0.9$. ``Wrong median'' uses the same cyclic within-layer wrong-$V$ control for every endpoint.}
\label{tab:closure-survey}
\begin{center}
\resizebox{\linewidth}{!}{%
\begin{tabular}{@{}lcrrrrrr@{}}
\toprule
Model & Topology & Heads & Median $\Pscore$ & q90 & Max $\Pscore$ & $\geq0.9$ & Wrong median \\
\midrule
Pythia-2.8B-deduped & MHA & 1024 & .4869 & .8827 & .9887 & 72 (7.03\%) & $1.37\!\times\!10^{-4}$ \\
Qwen3-4B Base & GQA & 1152 & .2752 & .8425 & .9799 & 67 (5.82\%) & $6.53\!\times\!10^{-5}$ \\
Mistral-7B-v0.3 & GQA & 1024 & .3236 & .8633 & .9855 & 72 (7.03\%) & $4.77\!\times\!10^{-5}$ \\
OLMo-2-13B, 5T & MHA & 1600 & .5251 & .8827 & .9854 & 128 (8.00\%) & $5.70\!\times\!10^{-5}$ \\
Qwen2.5-72B Base & GQA & 5120 & .2677 & .8036 & .9825 & 204 (3.98\%) & $7.14\!\times\!10^{-5}$ \\
Qwen3-235B-A22B & GQA & 6016 & .2794 & .8255 & .9894 & 338 (5.62\%) & $1.08\!\times\!10^{-4}$ \\
\bottomrule
\end{tabular}}
\end{center}
\end{table}

Because $\Pscore$ is scale-free, we assess return magnitude separately. For $\Pscore(T)>0$, define
\begin{equation}
 g_T=\frac{\lVert T^2\rVert_F}{\lVert T\rVert_F}
 =\frac{|\alpha^*|}{\sqrt{\Pscore(T)}}. \label{eq:return-gain}
\end{equation}
Across the 881 strong heads in Table~\ref{tab:closure-survey}, model-median $g_T$ ranges from 0.0317 to 2.512, its pooled minimum is 0.0144, and no strong head has $g_T<0.01$. Moreover, 880/881 strong heads exceed the model-specific all-head median return gain. The high-closure tail is therefore not driven by numerically square-zero maps; a modelwise quantile summary is reported in the supplement.

\subsection{Closure requires the head-specific pairing of $D$ and $K$}

Exhaustive reassignment gives a positive head-specific $D\!\times\!K$ interaction in all six families and 95.5--100\% of their layers. When $K,D$ are recomputed after wrong-$V$ pairing, the interaction collapses to approximately zero and the real-minus-control contrast is positive in 96.9--100\% of layers. This establishes pair specificity; the stricter intervention below isolates the responsible degree of freedom while fixing spectrum and support.

\subsection{Closure depends on within-support transport orientation}

Across all 7,304 heads, pooled median closure is 0.336 in the trained orientation and $1.04\times10^{-4}$ after scrambling. The trained orientation yields higher closure for 7,205/7,304 heads and a positive median difference in all 268 layers. The effect persists after excluding the 510 heads with $\Pscore\geq0.9$: 98.54\% of the remaining 6,794 heads still favor the trained orientation. None of the 58,432 matched counterfactuals reaches 0.85. Thus rank, energy, singular spectrum, read/write spans, and principal angles do not jointly determine closure. The largest counterfactual occurs in a near-rank-one head, consistent with the algebraic boundary that every non-nilpotent rank-one map closes automatically.

The basis-invariant Haar null gives the same population-level conclusion. The pooled Haar median is $1.67\times10^{-4}$, and trained closure exceeds the median of two Haar reorientations for 98.38\% of heads. The median paired difference is positive in all nine models and all 268 layers. Singular values are preserved to relative error below $10^{-12}$. The orientation effect is therefore not an artifact of choosing a particular basis inside the principal subspaces.

\begin{figure}[H]
\begin{center}
\includegraphics[width=0.98\linewidth]{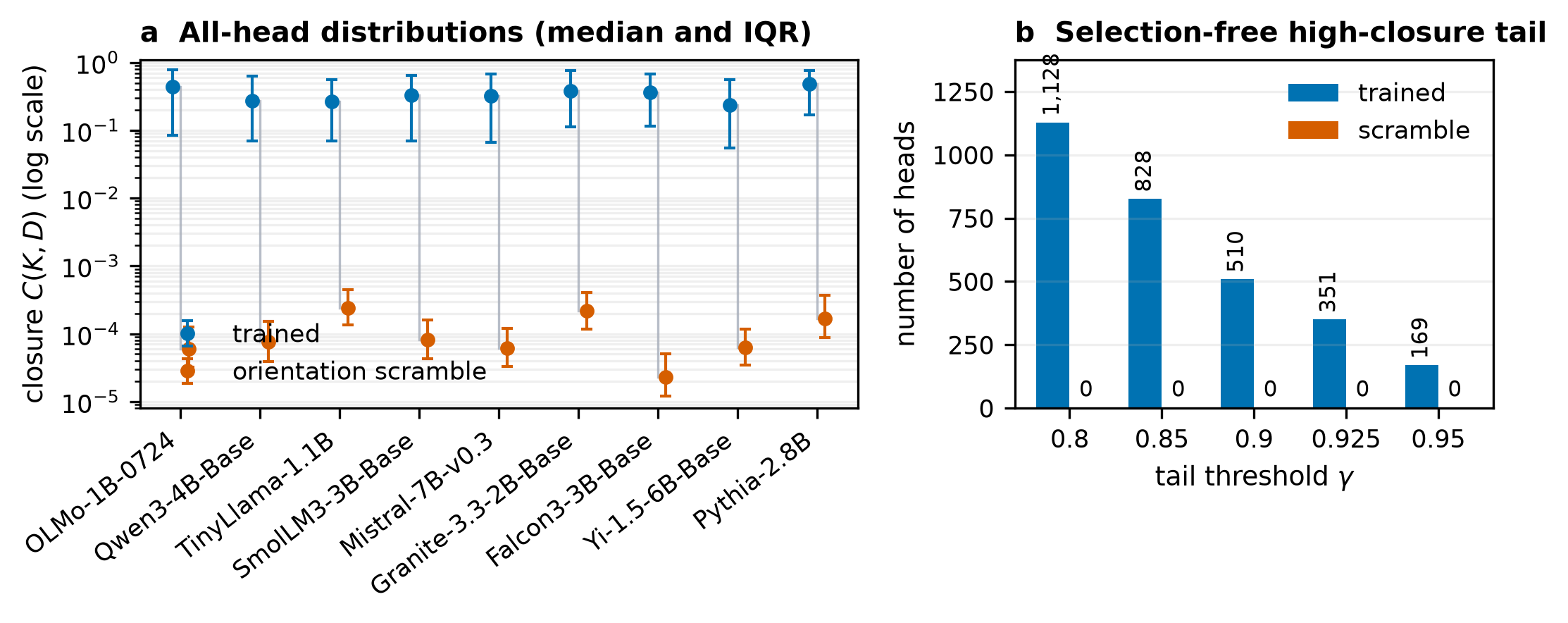}
\end{center}
\caption{A population-wide, spectrum- and support-preserving intervention removes closure throughout the head population. (A) Model medians and interquartile ranges for trained heads and the per-head median of eight matched transport-core scrambles. (B) Trained and fixed-first-scramble tail counts over all 7,304 heads; none of the 58,432 counterfactuals reaches 0.85.}
\label{fig:spectral-intervention}
\end{figure}

Constrained reorientation gives the complementary result. Across all nine models, the constructed feasible median is 0.988--0.994 for strong heads and 0.880--0.926 for layer-matched ordinary heads. The achieved lower-bound gap is only 11.5--20.8\% as large as the observed closure gap. Strong heads therefore realize a substantially larger fraction of the explicitly constructed feasible range; numerical-route and invariant-preservation audits support the calculation.

A sample chosen without reference to closure yields model-median feasible capacity of 0.884--0.927. All 268 layer medians exceed 0.8, including the 82 layers that contain no strong head. Among 291 sampled heads with feasible capacity of at least 0.9, 87.3\% nevertheless remain below the strong-head threshold. High capacity is therefore widespread but insufficient to explain which heads attain closure.

The retrospective training trajectories show the same constructed-capacity--attainment separation (Table~\ref{tab:capacity-dynamics}). OLMo and Pythia begin at initialization with nearly identical constructed capacity in the two endpoint-defined groups and no attainment-index gap; TinyLlama's first public checkpoint occurs after 10B training tokens. At the final checkpoint, the attainment-index gap is positive in all three models and all 54 eligible layers, with a hierarchical-bootstrap 95\% interval of 0.468--0.643. In every lineage, the final achieved lower-bound gap is less than half the observed closure gap, and the attainment gap grows more than that lower-bound gap. These trajectories establish a robust retrospective separation, not that the final labels or their causal origin were known at early checkpoints.

\begin{table}[H]
\caption{Constructed capacity and attainment across three training trajectories. Early columns give median feasible lower bounds for future strong (S) and matched ordinary (O) heads. Final columns are between-population median gaps. TinyLlama begins at 10B tokens; the other two begin at initialization.}
\label{tab:capacity-dynamics}
\centering
\small
\begin{tabular}{lccccc}
\toprule
Model & Early $\widehat{\mathrm{Cap}}_S$ & Early $\widehat{\mathrm{Cap}}_O$ & Final $\Delta\Pscore$ & Final $\Delta\widehat{\mathrm{Cap}}$ & Final $\Delta E$ \\
\midrule
OLMo-1B & 0.839 & 0.839 & 0.412 & 0.083 & 0.353 \\
TinyLlama-1.1B & 0.894 & 0.899 & 0.536 & 0.111 & 0.488 \\
Pythia-2.8B-deduped & 0.806 & 0.805 & 0.368 & 0.069 & 0.321 \\
\bottomrule
\end{tabular}
\end{table}

\begin{figure}[H]
\begin{center}
\includegraphics[width=0.98\linewidth]{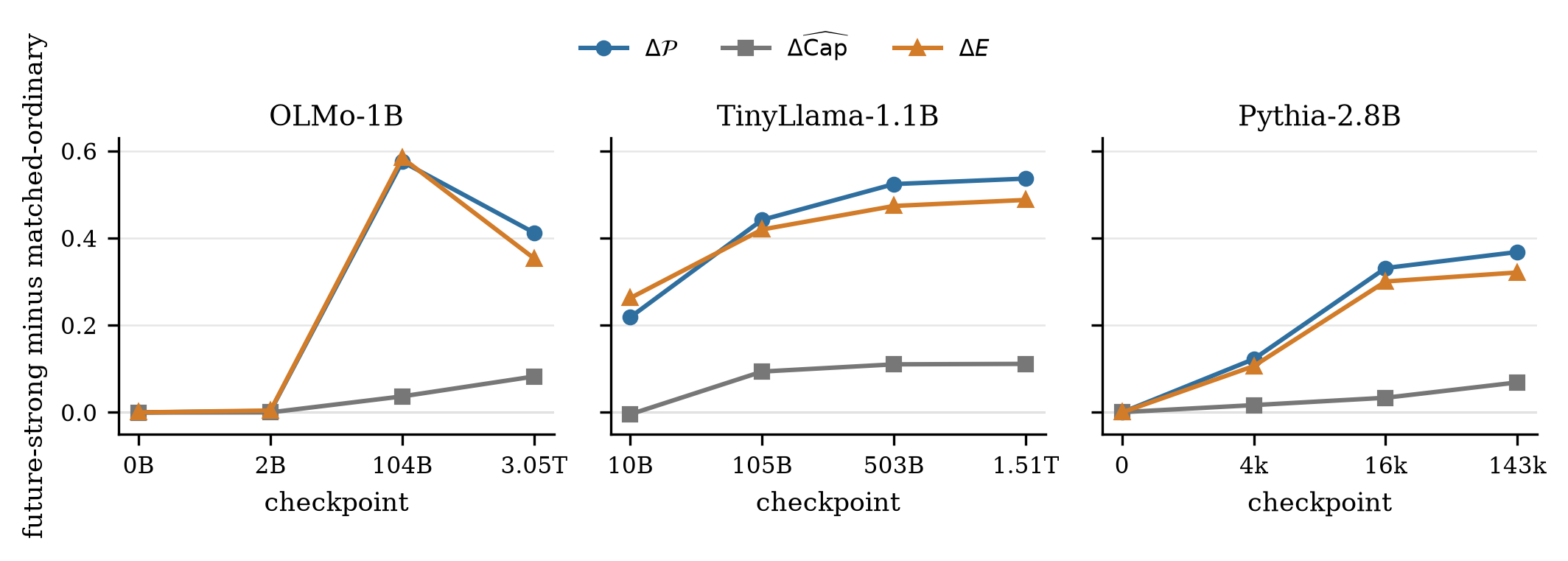}
\end{center}
\caption{Complete saved-checkpoint trajectories for the endpoint-defined populations. Each curve is the future-strong minus layer-matched-ordinary median gap in observed closure $\Pscore$, constructive capacity $\widehat{\operatorname{Cap}}$, or constructed-range attainment index $E$. Horizontal spacing indexes the four saved checkpoints rather than elapsed training time; TinyLlama begins at its first public 10B-token checkpoint. The lower-bound gap remains smaller while the attained-orientation gap carries most of the later separation.}
\label{fig:capacity-dynamics-full}
\end{figure}

These are constructive capacity--attainment separations, not claims that the numerical search finds the global optimum.

\subsection{Supporting check on natural activations}

Across OLMo-1B, TinyLlama-1.1B, SmolLM3-3B, and Qwen3-4B-Instruct, the paired return preserves the direction of natural outputs for high-closure heads (model medians 0.842--0.886), whereas wrong-$V$ returns are below $10^{-3}$ and layer-matched low-closure heads are near 0.012. Both contrasts are positive in all 86 eligible layers. The corresponding median RMS return magnitudes are 0.088, 2.071, 0.412, and 1.239, compared with 0.038, 0.867, 0.146, and 0.432 for layer-matched low-closure heads. Thus the directional result is not produced by vanishing returns on the sampled activations. This remains a data-weighted consistency check rather than evidence that the forward pass executes or behaviorally requires a second application.

\subsection{Scale-free endpoint geometry of the strong tail}

Isoclinic normalization exposes the general core law from Equation~\ref{eq:latent-closure}. Across twelve endpoints, 1,088/1,180 strong heads remain above 0.9 after replacing trained support by $D_{\mathrm{iso}}$; the pooled latent score is 0.957. This is not a generic low-rank effect: over all 7,304 local heads, latent closure separates the trained strong tail from ordinary heads with pooled AUROC 0.984, and a layer-matched 13B--235B additional-model extension gives the same pooled AUROC.

Support shape explains much of the remaining variation. Conditional on latent closure, $I_D$ distinguishes cores whose closure is preserved from those whose closure is reduced, with pooled AUROC 0.945 in the local models and 0.963 in the large-endpoint extension. Both separations remain strong throughout the common 0.80--0.95 threshold sweep and under model/layer-clustered inference. This yields an endpoint decomposition in which most strong heads contain a latent $K^2\approx\beta K$ core, while trained support determines how faithfully that relation is preserved in $KDK$.

Equation~\ref{eq:perturbation-identity} identifies the corresponding mechanism. Cores with reduced closure have larger nonradial $KEK$ responses in every local family; off-plane leakage accounts for most of the reduction and can only decrease closure. For a smaller subset, closure also depends on how the same principal-angle spectrum is assigned to transport axes. The perturbation bound and axis-permutation controls are reported in the supplement.

\subsection{Shared values extend self-closure into fixed-gain right action}

Proposition~\ref{prop:shared-value-algebra} predicts that the headwise closure law extends to siblings sharing the same value factor. We evaluate this prediction for 406 strong right heads and 1,458 sibling compositions (Table~\ref{tab:gqa-algebra}). Without refitting the coefficient, the right head's own $\alpha_j$ explains 0.913--0.936 of median sibling-product energy. Freely fitted sibling gains satisfy $\rho_{ij}=\beta_{ij}/\alpha_j$ with model medians of 0.959--0.974, whereas matched cross-value controls retain only 0.013--0.053 of that gain.

The additional Yi-1.5-6B evaluation reproduces the separation after the coefficient and control definitions were fixed: all 20 informative layers favor the shared-value law, with median fixed-gain explanation of 0.922 and a gain ratio of 0.963. Thus both direction and signed magnitude are inherited through the shared value channel.

\begin{table}[H]
\caption{Gain-preserving shared-$V$ algebra. $F(\alpha_j)$ fixes the coefficient to the right head's self-fitted gain; $\rho=\beta_{ij}/\alpha_j$ reports the freely fitted coefficient relative to that gain. Control uses the same query slot in the next KV group. $^*$Yi-1.5-6B was evaluated after the coefficient and control definitions were fixed.}
\label{tab:gqa-algebra}
\begin{center}
\small
\resizebox{\linewidth}{!}{%
\begin{tabular}{lrrrrrr}
\toprule
Model & Right heads & Pairs & Layers & $F(\alpha_j)$ & Same $\rho$ & Control $\rho$ \\
\midrule
TinyLlama-1.1B & 32 & 224 & 14 & .932 & .974 & .013 \\
SmolLM3-3B & 35 & 105 & 18 & .915 & .966 & .019 \\
Qwen3-4B Base & 67 & 201 & 21 & .928 & .964 & .027 \\
Mistral-7B & 72 & 216 & 24 & .934 & .970 & .014 \\
Granite-3.3-2B & 142 & 426 & 34 & .936 & .966 & .017 \\
Falcon3-3B & 24 & 48 & 15 & .913 & .959 & .053 \\
Yi-1.5-6B$^*$ & 34 & 238 & 20 & .922 & .963 & .022 \\
\bottomrule
\end{tabular}}
\end{center}
\end{table}

The operators remain distinct despite obeying a common product law. For normalized strong heads $P_h=T_h/\alpha_h$, 157 unordered sibling pairs (314 directed products) satisfy the approximate left-zero-band relation $P_iP_j\approx P_i$ and $P_jP_i\approx P_j$. In the five families with at least ten such pairs, median operator squared cosine is 0.315--0.486, whereas median fixed-scale product explanation is 0.935--0.952. At least 98.6\% of pairs remain below 0.8 squared cosine.

Equation~\ref{eq:shared-value-algebra} explains this nonduplication. For each strong GQA head, we compare $P_h$ with the canonical anchor $P_V=V(V^\top V)^{-1}V^\top$. Across all seven families, the model-median squared cosine is 0.449--0.639, while 0.373--0.553 of operator energy lies in the free oblique component. Every strong head favors its trained anchor over wrong-$V$, and exhaustive same-layer retrieval ranks the trained $V$ first for all 510 strong heads in the nine locally complete models. The empirical band is therefore the approximate counterpart of $\mathcal A_V$: a common value-defined kernel coexists with distinct write-side degrees of freedom.

The additional-model scale extension confirms the anchor geometry for all 670 strong heads in OLMo-2-13B, Qwen2.5-72B Base, and Qwen3-235B-A22B. Together with the seven-family analysis, the result covers 1,076 strong heads across ten endpoints from 1.1B to 235B parameters. MHA confirmation shows that the anchor is intrinsic to paired O/V organization, while GQA sharing supplies the cross-head product law.

\section{Discussion}

\subsection{What $T^2\approx\alpha T$ encodes}

The $KDK$ decomposition and matched orientation intervention show that scaled idempotence is carried by a specific relation between separately parameterized but jointly trained read and write factors. Spectrum, rank, support, and principal angles can all remain fixed while closure disappears. The resulting object is an oriented return geometry rather than a generic consequence of low rank or a token-copying map. The negative fitted coefficient found in 69--76\% of large-endpoint strong heads further supports this signed interpretation, while the return-gain analysis excludes a numerically square-zero explanation.

This geometry is widely available but sparsely attained. Heads sampled independently of closure can be reoriented toward high closure in every surveyed layer, including layers in which the trained endpoint contains no strong head. In the three retrospective trajectories, the endpoint-defined groups separate primarily in attained orientation rather than feasible capacity. The sparse upper tail is therefore associated with the orientation present at the trained endpoint inside a large feasible set. Our experiments characterize that separation geometrically; identifying the data, gradients, and downstream objectives that produce it remains an open question.

The latent-core decomposition provides a complementary view. Most strong heads satisfy $K^2\approx\beta K$ after isoclinic normalization. Trained support then determines how this core appears in $KDK$: radial response preserves projective closure, whereas nonradial response rotates the operator or leaks energy away from its closure line. This decomposition describes the endpoint geometry of the observed weights but does not by itself identify the training process that produced them.

Value sharing turns the headwise relation into an algebraic one. Proposition~\ref{prop:shared-value-algebra} shows that an exactly closed right head transfers its coefficient to every sibling with the same value factor. The algebra permits distinct oblique projections with a common value-defined kernel, so the heads need not be duplicates. Seven-model fixed-gain tests and the ten-endpoint anchor analysis place pretrained weights near this exact solution family. In GQA, shared values therefore define the domain on which a headwise self-relation becomes a cross-head product law.

\subsection{Limitations and future work}

The present conclusions are structural. The natural-activation analysis shows that the diagnostic return survives sampled hidden-state covariance, but does not establish that the forward pass explicitly composes these operators or that language-model predictions depend on closure. The three trajectories use endpoint-conditioned groups and describe capacity and attainment rather than the causal dynamics of optimization. The cross-head product law requires exact value sharing and does not directly extend to MHA. Establishing functional relevance will require interventions in the full context-conditioned path that control first-pass activation changes, QK feedback, normalization, and surrounding computation.

\subsubsection*{Broader Impact Statement}

Understanding recurrent internal transport geometry may support model auditing, debugging, and controlled modification. The same insight could enable more targeted manipulation of learned behavior; applications should therefore distinguish structural diagnostics from claims about semantics or capability.

\section{Conclusion}

Transformer attention repeatedly contains OV operators that nearly satisfy $T^2\approx\alpha T$. Their closure is explained by an exact $KDK$ return geometry and depends on trained transport orientation, not on low rank, spectrum, support, principal angles, or a near-zero return alone. High closure is constructively feasible throughout the network but attained only by a sparse population, separating geometric capacity from the trained endpoint. Under shared values, the same self-relation extends to the fixed-gain cross-head law $T_iT_j\approx\alpha_jT_i$, producing distinct oblique operators with a common value-defined kernel. These results identify scaled idempotence as a recurrent geometry in trained weights and show how it induces a local operator algebra in Transformer attention.

\subsubsection*{Author Contributions}

Jiming Feng led the conception, experimental design, implementation, analysis, and manuscript preparation. Junliang Li contributed through research discussions and experimental development.

\subsubsection*{Reproducibility Statement}

A reproducibility package has been prepared for this work. All figures are generated from fixed result artifacts; the package contains archival analysis entry points, recorded model identifiers and revisions, selection rules, seeds, controls, per-head outputs, and automated checks of the retained numerical claims. No model weights are redistributed.

\begingroup
\small
\bibliography{main}
\bibliographystyle{tmlr}
\endgroup

\end{document}